\documentclass[10pt,twocolumn,letterpaper]{article}

\usepackage[final]{cvpr}      
\definecolor{cvprblue}{rgb}{0.21,0.49,0.74}
\usepackage[pagebackref,breaklinks,colorlinks,allcolors=cvprblue]{hyperref}

\usepackage{soul}
\usepackage[utf8]{inputenc} 
\usepackage[T1]{fontenc}    
\usepackage{hyperref}       
\usepackage{url}            
\usepackage{booktabs}       
\usepackage{amsfonts}       
\usepackage{nicefrac}       
\usepackage{microtype}      
\usepackage{xcolor}         
\usepackage{multirow}
\usepackage{colortbl}
\usepackage{booktabs}
\usepackage{graphicx}
\usepackage{amsmath}
\usepackage{bm}
\usepackage{amssymb}
\usepackage{graphicx,wrapfig,lipsum}
\usepackage{caption}
\usepackage{tabularx}
\usepackage{bbding}
\usepackage{placeins} 
\usepackage{dsfont}
\usepackage[accsupp]{axessibility}  

\def\paperID{10699} 
\def\confName{CVPR}
\def\confYear{2026}

\title{Flow4DGS-SLAM: Optical Flow-Guided 4D Gaussian Splatting SLAM}

\author{%
    Yunsong Wang  \qquad Gim Hee Lee \\
    School of Computing, National University of Singapore \\ 
    \texttt{yunsong@comp.nus.edu.sg} \quad
    \texttt{gimhee.lee@nus.edu.sg}\\
    {\tt \href{https://wangys16.github.io/Flow4DGS-SLAM}{\textbf{https://wangys16.github.io/Flow4DGS-SLAM}}}
}

\begin{document}

\twocolumn[{%
\renewcommand\twocolumn[1][]{#1}%
\maketitle
\begin{center}
    \centering
    \captionsetup{type=figure}
    \includegraphics[width=1\textwidth]{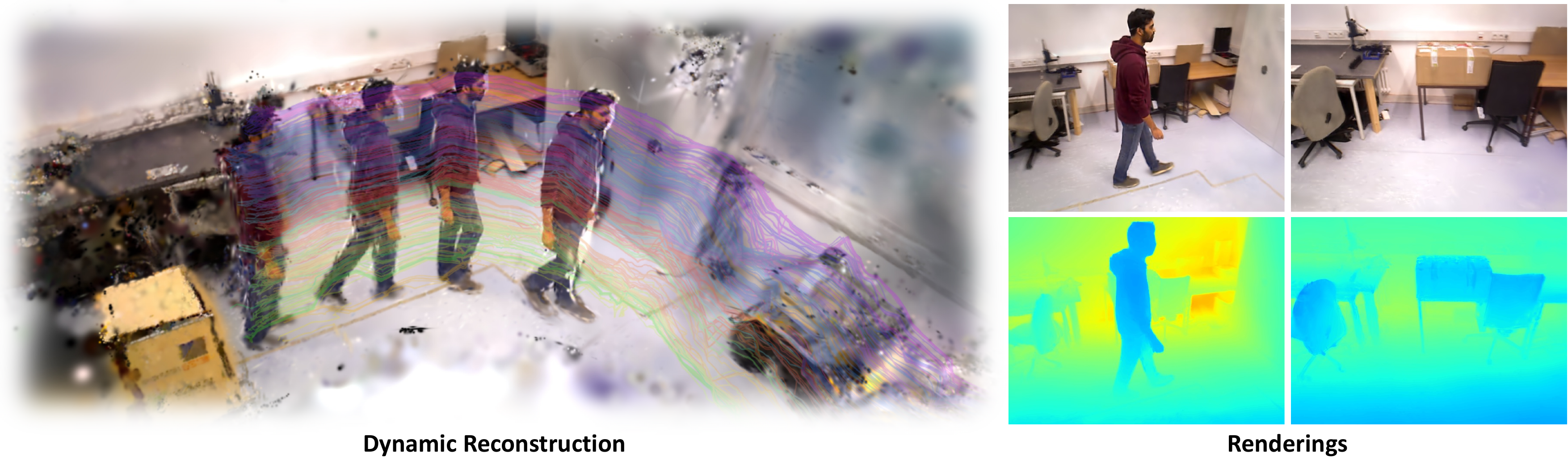}
    \captionof{figure}{\textbf{Overview of our results.} Our method achieves high-quality renderings with spatially and temporally coherent Gaussian motion.}
    \label{teaser}
\end{center}%
}]
\maketitle

\begin{abstract}
    Handling the dynamic environments is a significant research challenge in Visual Simultaneous Localization and Mapping (SLAM). Recent research combines 3D Gaussian Splatting (3DGS) with SLAM to achieve both robust camera pose estimation and photorealistic renderings. However, using SLAM to efficiently reconstruct both static and dynamic regions remains challenging. In this work, we propose an efficient framework for dynamic 3DGS SLAM guided by optical flow. Using the input depth and prior optical flow, we first propose a category-agnostic motion mask generation strategy by fitting a camera ego-motion model to decompose the optical flow. This module separates dynamic and static Gaussians and simultaneously provides flow-guided camera pose initialization. We boost the training speed of dynamic 3DGS by explicitly modeling their temporal centers at keyframes. These centers are propagated using 3D scene flow priors and are dynamically initialized with an adaptive insertion strategy. Alongside this, we model the temporal opacity and rotation using a Gaussian Mixture Model (GMM) to adaptively learn the complex dynamics. The empirical results demonstrate our state-of-the-art performance in tracking, dynamic reconstruction, and training efficiency.
\end{abstract}

\section{Introduction}

Visual Simultaneous Localization and Mapping (SLAM) remains a fundamental yet challenging research task. It jointly estimates camera motion and builds a 3D scene map to enable applications such as robotics, AR/VR, and autonomous systems. However, most existing systems still assume a static world~\cite{orbslam2,orbslam3,imap,nice_slam,co-slam,eslam,voxfusion}. In real-world dynamic environments, this assumption often causes drift or tracking failure. 
Recently, 3D Gaussian Splatting (3DGS)~\cite{3dgs} has been integrated into SLAM to achieve real-time, photorealistic rendering and explicit 3D mapping~\cite{yan2023gs,keetha2023splatam,monogs,yugay2023gaussianslam}. These methods improve convergence in pose optimization and support online mapping, but typically treat dynamic objects as outliers, removing them to stabilize tracking~\cite{dg-slam,kong2024dgs,wildgsslam}. Therefore, the final maps capture only the static backgrounds.

Dynamic reconstruction studies have extended 3DGS to the temporal domain by learning MLP-based deformation field~\cite{deformable, scgs, liang2025gaufre, guo2024motion}, parameterizing Gaussian motion~\cite{spacetime, shapeofmotion}, or directly modeling temporal offsets~\cite{duan20244d, marbles, yang2023real}. However, these approaches typically rely on precomputed multi-view poses and lengthy offline training. Therefore, combining dynamic 3DGS with SLAM offers great potential for joint camera tracking and efficient dynamic reconstruction.

One prior work, 4DGS-SLAM~\cite{4dgs-slam}, represents an initial step in this direction. It leverages sparse control points from SC-GS~\cite{scgs} to reconstruct dynamic elements, and estimates camera motion using the static 3DGS. However, its deformation field training is computationally expensive, hindering real-time performance. Moreover, its category-based segmentation model struggles in more general dynamic scenes, and its reconstruction capability is limited in handling complex dynamics (\textit{e.g.}, people leaving and re-entering the view). To address these challenges, we propose \textbf{Flow4DGS-SLAM}, a dynamic SLAM framework guided by optical flow for efficient tracking and reconstruction in complex dynamic environments.

To handle more general dynamic scenes, we first design a Camera-Induced Motion Decomposition module to perform category-agnostic motion segmentation. This module leverages the input depth map and prior optical flow to efficiently solve for the camera ego-motion, estimating the rigid flow induced by camera motion and filter out the outlier dynamic pixels. 
We further represent the dynamic Gaussians in a hybrid form, with explicitly learned Gaussian centers at keyframes and parametric Gaussian-Mixture Model (GMM)-based temporal opacity and rotation. 
With the explicit Gaussian centers, we take inspiration from GFlow \cite{gflow} to leverage the off-the-shelf optical flow models for explicit propagation to efficiently model dynamics of the Gaussians, which is combined with a KNN-based smoothing strategy to ensure local rigidity. 
Furthermore, we leverage optical flow to back-track newly appearing motion regions, in order to adaptively insert new dynamic Gaussians. 
Additionally, the GMM-based temporal opacity and rotation improves the capability to reconstruct complex dynamic scenes without significantly enlarging the model size. 

Our \textbf{contributions} can be summarized as follows:
\begin{enumerate}
    \item \textbf{Camera-Induced Motion Decomposition.} 
    We design a parametric model that efficiently generates a category-agnostic motion mask to handle more general dynamic scenes. The model fits a 6-DoF camera motion to the depth map and optical flow, providing a better camera initialization and filtering out the outlier dynamic pixels.
    \item \textbf{Hybrid 4DGS guided by optical flow.} We propose a hybrid 4D Gaussian representation with explicit temporal positions and GMM-based opacity and rotation. The explicit position modeling is guided by a scene-flow Gaussian propagation module and an adaptive gaussian insertion module to accelerate training. 
    The GMM-based opacity and rotation improves the 
    representation capability of the model to reconstruct complex dynamic scenes.
    \item \textbf{Superior Performance.} The empirical results demonstrate our state-of-the-art performance in tracking, photorealistic rendering, and computational efficiency.
\end{enumerate}

\begin{figure*}
    \centering
    \includegraphics[width=\linewidth]{images/method/framework.pdf}
    \caption{\textbf{Overall framework of Flow4DGS-SLAM.} Given input RGB-D video, we first extract the prior semantic mask and optical flow, and feed them into a camera-induced motion decomposition module to filter out category-agnostic motion mask and solve an optical-flow guided camera initialization. The static gaussians help refine the camera pose during tracking, and the dynamic Gaussians are represented in a hybrid form, combined with a scene flow Gaussian propagation module and an adaptive gaussian insertion module to accelerate training.}
    \label{fig:framework}
\end{figure*}
\section{Related Work}

\noindent\textbf{Static SLAM.}
Classic visual SLAM decomposes tracking and mapping, relying on sparse keypoints to estimate camera motion~\cite{orbslam2,orbslam3}. Dense SLAM extends this by reconstructing per-frame geometry from depth maps anchored to keyframes~\cite{newcombe2011dtam,ummenhofer2017demon,zhou2018deeptam,teed2018deepv2d}. The frame-centric methods \cite{czarnowski2020deepfactors, teed2021droid} efficiently estimate frame-to-frame camera motion, but may face inconsistencies when assembling into a global map. In contrast, map-centric methods use a unified 3D map representation, such as surfels~\cite{whelan2015elasticfusion,schops2019bad} or voxel-based TSDF grids~\cite{curless1996volumetric,newcombe2011kinectfusion,dai2017bundlefusion}. 
More recent works leverage NeRF or 3DGS~\cite{imap,nice_slam,co-slam,keetha2023splatam,monogs} via differentiable rendering for photorealistic tracking and mapping. Although effective in static environments, these systems often fail in dynamic scenes where moving objects significantly hinder tracking.

\vspace{1pt}
\noindent\textbf{Gaussian Splatting.}
3D Gaussian Splatting (3DGS)~\cite{3dgs} achieves real-time rendering by optimizing anisotropic Gaussians, showing strong performance in surface reconstruction~\cite{sugar,neusg,vcr}, feed-forward reconstruction~\cite{pixelsplat,mvsplat,freesplat,freesplat++}, and SLAM~\cite{monogs,keetha2023splatam}. To handle dynamic scenes, some subsequent works learn temporal deformation fields through MLPs~\cite{deformable,scgs,liang2025gaufre,guo2024motion}, leveraging the low-frequency bias of MLPs for smooth temporal and spatial deformation. For instance, SC-GS~\cite{scgs} trains sparse control points and an MLP-based deformation field where Gaussian motion is computed based on weighted sum of its neighboring control points' motion, ensuring spatial rigidity and temporal coherence. However, such approaches require hours of optimization due to costly deformation MLPs training. Others optimize Gaussian trajectories parametrically~\cite{spacetime,shapeofmotion} or directly learn temporal Gaussian parameters~\cite{luiten2024dynamic,yang2023real}. Since dynamic SLAM requires efficient reconstruction from sparse keyframes, we propose a hybrid dynamic Gaussian representation that combines the efficiency of explicit modeling with the temporal coherence of parametric approaches.

\vspace{1pt}
\noindent\textbf{SLAM in Dynamic Scenes.}
Typical SLAM systems handle dynamics by detecting and excluding moving regions using geometric residuals or semantic priors~\cite{ds_slam,dynaslam,vdoslam,dg-slam,wildgsslam}. For example, WildGS-SLAM~\cite{wildgsslam} employs DINO~\cite{dino} features and an uncertainty MLP to distinguish dynamic from static regions, stabilizing tracking while yielding only static background maps. ADD-SLAM \cite{wu2025add} reconstructs both the static and dynamic regions on keyframes, but lacks deformation of the dynamic Gaussians for temporal view interpolation. 4DGS-SLAM~\cite{4dgs-slam} advances this by reconstructing dynamic Gaussians online via a sparse-control-point deformation field, but faces intensive computational cost and becomes less effective in more general and complex dynamic scenes. In contrast, our method explicitly models Gaussian motion over time with scene-flow priors, enabling fast and robust reconstruction of both static and dynamic elements within the SLAM pipeline.

\section{Our Methodology}

\noindent \textbf{Overview.} As shown in Figure \ref{fig:framework}, our framework performs online tracking of camera motion and mapping of a 4D scene map. For tracking, we propose a Camera-Induced Motion Decomposition approach to generate a category-agnostic motion mask, which is combined with the semantic mask generated by YOLOv9 \cite{wang2024yolov9} to avoid dynamic distractors during tracking. For dynamic mapping, we represent dynamic Gaussians in a hybrid form with explicit temporal positions. This allows us to leverage prior optical flow for explicit Gaussian propagation and adaptive Gaussian insertion. In parallel, the time-varying opacities and rotations of dynamic Gaussians are modeled with a Gaussian Mixture Model for the reconstruction of more complex dynamic scenes.

\subsection{Camera-Induced Motion Decomposition}
\label{sec:motion}

\textbf{Category-agnostic Masking.} 
We leverage optical flow and depth to fit a camera ego-motion model. This model predicts the rigid flow induced by camera motion, which allows us to identify and filter out dynamic pixels. The estimated camera motion also provides a coarse initialization of the new camera pose.
Specifically, given the new RGB-D frame $\{\mathbf{I}^t, \mathbf{D}^t\}$ at timestamp $t$, we first feed $\mathbf{I}^t$ and $\mathbf{I}^{t-1}$ into RAFT~\cite{raft} to obtain optical flow $\mathbf{F}^{t, t-1}(u,v)$ 
which we denote as $\mathbf{F}(u,v)$ for brevity. Let $Z=\mathbf{D}^t(u,v)$ and $\bm{x}=(u,v,Z)^\top$ denote the corresponding 3D point. Under small camera motion from $t$ to $t\!-\!1$, the motion field of a static 3D point \cite{horn1981determining} is:
\begin{equation}
\label{eq:motion_field}
\mathbf{F}(u,v)=\mathbf{J}(\bm{x})\,\boldsymbol{\xi},
\end{equation}
where the camera twist $\boldsymbol{\xi}=[{\bm{\rho}}^\top,{\bm{\theta}}^\top]^\top\!\in\!\mathbb{R}^6$ contains translation and rotation. The $2{\times}6$ image Jacobian $\mathbf{J}$ follows projective geometry~\cite{chaumette_visualservoing}:
\begin{equation}
\label{eq:L_matrix}
\mathbf{J}(\bm{x})
=
\begin{bmatrix}
-\frac{f_x}{Z} & 0 & \frac{u}{Z} &
\frac{uv}{f_y} &
-f_x-\frac{u^2}{f_x} &
v\\[2pt]
0 & -\frac{f_y}{Z} & \frac{v}{Z} &
f_y+\frac{v^2}{f_y} &
-\frac{uv}{f_x} &
-u
\end{bmatrix},
\end{equation}
where $(f_x, f_y)$ is the camera intrinsics for brevity. The translation-induced flow scales with $1/Z$, while rotation-induced flow is depth-independent. We further obtain a semantic mask $\mathcal{M}_s$ from YOLOv9~\cite{wang2024yolov9}, where $\mathcal{M}_s(u,v)=1$ indicates dynamic objects (people). Assuming that most pixels with $\mathcal{M}_s(u,v)=0$ are static, we stack Eq. \eqref{eq:motion_field} over all pixels $(u,v)$ with $Z(u,v)>0$ and $\mathcal{M}_s(u,v)=0$, and solve:
\begin{equation}
\label{eq:wls}
\hat{\boldsymbol{\xi}}
=
\arg\min_{\boldsymbol{\xi}}
\sum_{i} w_i\,\|\mathbf{F}_i-\mathbf{J}_i\boldsymbol{\xi}\|^2,
\end{equation}
using IRLS with Cauchy weights $\{w_i\}$. The predicted rigid flow is $\hat{\mathbf{F}}(u,v)=\mathbf{J}(u,v,Z)\hat{\boldsymbol{\xi}}$. We compute residuals:
\begin{equation}
\label{eq:residual}
r(u,v)=\|\mathbf{F}(u,v)-\hat{\mathbf{F}}(u,v)\|_2.
\end{equation}
Since 
dynamic pixels tend to yield large residuals, we obtain the category-agnostic mask using the threshold: 
\begin{subequations}
\label{eq:mask}
\small
\begin{align}
&\mathcal{M}_{ca}(u,v)
=
\mathds{1}\!\left(
r(u,v)>
\mathrm{median}(r)+k\,\mathrm{MAD}(r)\right),
\\
&\text{where} \qquad
\mathrm{MAD}(r)
=
\mathrm{median}_i\bigl|r_i-\mathrm{median}_j(r_j)\bigr|.
\end{align}
\end{subequations}
The final dynamic mask is $\mathcal{M}_{dy}=\mathcal{M}_s\cup\mathcal{M}_{ca}$.

\vspace{3pt}\noindent\textbf{Camera Initialization.}
To solve for a more accurate camera motion, we use pixels with $\mathcal{M}_{dy}(u,v)=0$ and valid depth to perform a second weighted least-squares refinement 
for a cleaner twist $\hat{\boldsymbol{\xi}}^{\!*}$. We map $\hat{\boldsymbol{\xi}}^{\!*}$ to $\mathrm{SE}(3)$ via the exponential map:
\begin{equation}
\label{eq:se3_exp}
\exp_{\mathfrak{se}(3)}(\hat{\boldsymbol{\xi}}^{\!*})
=
\begin{bmatrix}
\exp_{\mathfrak{so}(3)}(\bm{\theta}^{\!*}) &
\mathbf{V}(\bm{\theta}^{\!*})\,\bm{\rho}^{\!*} \\
\mathbf{0}^\top & 1
\end{bmatrix},
\end{equation}
where $\mathbf{V}(\cdot)$ is the left Jacobian of $\mathrm{SO}(3)$. With previous pose
$\mathbf{T}_{cw}^{t-1}=
\big[\mathbf{R}^{t-1}\mid\mathbf{t}^{t-1}\big]$,
we initialize the current pose as:
\begin{equation}
\label{eq:pose_update_final}
\mathbf{T}^t_{cw}
=
\mathbf{T}^{t-1}_{cw}\;
\exp_{\mathfrak{se}(3)}(\hat{\boldsymbol{\xi}}^{\!*}).
\end{equation}
In experiments, we clamp the maximum camera motion with a threshold proportional to the ratio of inlier pixels to improve the robustness to optical flow noise. 
Our camera-induced motion decomposition leverages prior optical flow and the image Jacobian to efficiently generate a category-agnostic motion mask and provide a better camera pose initialization. As a result, the tracking robustness of dynamic SLAM is significantly improved in complex dynamic scenes.

\subsection{Hybrid 4D Gaussian Splatting Representation}

\vspace{3pt}\noindent\textbf{Preliminary of 3DGS.} Each static 3D Gaussian $\bm{\mathcal{G}}_i^s$ is defined by a center position 
$\mathbf{x}_i^s \in \mathbb{R}^3$, a covariance matrix 
$\mathbf{\Sigma}_i^s \in \mathbb{R}^{3 \times 3}$, 
an opacity $\sigma_i^s \in [0,1]$, and color parameters $\bm{c}_i^s$. 
During rendering, each Gaussian is projected onto the image plane and accumulated in a back-to-front order using an $\alpha$-blending scheme. The rendered color map is:
\begin{equation}
    \label{eq:render}
    \hat{\mathbf{C}}^s(\mathbf{u}) = 
    \sum_{i=1}^{|\mathcal{G}|}
        \bm{c}_i^s\,\alpha_i^s(\mathbf{u}) 
        \prod_{j<i} (1 - \alpha_j^s(\mathbf{u})),
\end{equation}
where $\alpha_i^s(\mathbf{u})$ represents the elliptical footprint of $\bm{\mathcal{G}}_i^s$ at pixel $\mathbf{u}$ multiplied by $\sigma_i^s$. 
The depth map $\hat{\mathbf{D}}^s(\mathbf{u})$ and opacity map $\hat{\mathbf{O}}^s(\mathbf{u})$ are rendered similarly by replacing $\bm{c}_i^s$ in Eq. \eqref{eq:render} with depth and 1, respectively.

\vspace{3pt}\noindent\textbf{Hybrid 4DGS.} We separate the 3D Gaussians into static and dynamic, which are defined and modeled differently. The static Gaussians are defined similarly as in 3DGS \cite{3dgs}. 
We define the 4D Gaussian Splatting in a hybrid form, including explicit time-varying trajectories and Gaussian Mixture Model (GMM)-based temporal opacities and rotations. The $i$-th dynamic Gaussian is defined by a set of static attributes $\{\bm{s}_i, \sigma_i, \bm{c}_i\}$, where $\bm{s}_i$ is its scale, $\sigma_i$ is its static opacity, and $\bm{c}_i$ denotes color features. 
The dynamic attributes consist of: i) explicit positions $\mathbf{x}_i(t)$ at keyframes, and ii) temporally continuous opacity $\sigma_i(t)$ and rotation $\mathbf{q}_i(t)$.

\smallskip\noindent\textbf{Explicit Keyframe Positions.} To accelerate 
training, guided by the prior optical flow introduced in Section \ref{sec:mapping}, we first discretize the temporal domain into a small number of keyframes $\{t_k\}$. 
For the $i$-th dynamic Gaussian, we learn its 3D center $\mathbf{x}_{i}^k$ at each keyframe. 
At arbitrary time $t$, the position is obtained via linear interpolation, which supports flexible motion and explicit operation for efficient training. 

\begin{table*}[t]
\small
\centering
\renewcommand{\arraystretch}{1} 
\setlength{\tabcolsep}{10pt}       
\setlength{\aboverulesep}{0pt}
\setlength{\belowrulesep}{0pt}
\setlength{\extrarowheight}{0pt}
\caption{\textbf{Trajectory ATE RMSE [cm]$\downarrow$ on the TUM RGB-D sequences.}
Best results are shown in \textbf{bold}.}
\vspace{-3mm}
\begin{tabular}{l|ccccccc}
\toprule
Method  & fr3/sit\_st & fr3/sit\_xyz & fr3/sit\_rpy & fr3/walk\_st & fr3/walk\_xyz & fr3/walk\_rpy & Avg. \\
\midrule
RoDyn-SLAM~\cite{jiang2024rodynslam}      & 1.5  & 5.6  & 5.7  & 1.7   & 8.3   & 8.1   & 5.1  \\
MonoGS~\cite{monogs} & \textbf{0.48} & 1.7 & 6.1  & 21.9  & 30.7  & 34.2  & 15.8 \\
SplaTAM~\cite{keetha2023splatam}          & 0.52 & 1.5  & 11.8 & 83.2  & 134.2 & 142.3 & 62.2 \\
4DGS-SLAM~\cite{4dgs-slam}                & 0.58 & 2.9  & 2.6  & 0.61  & 2.7   & \textbf{3.0} & 2.1 \\
\midrule
Ours                                      & 0.70 & \textbf{1.8} & \textbf{2.4} & \textbf{0.48} & \textbf{2.5} & 3.6 & \textbf{1.9} \\
\bottomrule
\end{tabular}
\label{tab:tum_ate}
\end{table*}

\begin{table*}[ht]
\small
\centering
\captionsetup[table*]{singlelinecheck=off}
\renewcommand{\arraystretch}{1} 
\setlength{\tabcolsep}{6.45pt}       
\setlength{\aboverulesep}{0pt}
\setlength{\belowrulesep}{0pt}
\setlength{\extrarowheight}{0pt}
\caption{\textbf{Quantitative results on the TUM RGB-D sequences.} 
Best results are highlighted in \textbf{bold}.}
\vspace{-3mm}
\begin{tabular}{l|c|ccccccc}
\toprule
Method & Metric 
& fr3/sit\_st & fr3/sit\_xyz & fr3/sit\_rpy 
& fr3/walk\_st & fr3/walk\_xyz & fr3/walk\_rpy & Avg. \\
\midrule
\multirow{3}{*}{MonoGS \cite{monogs}} 
& PSNR{[}dB{]}$\uparrow$  & 19.95 & 23.92 & 16.99 & 16.47 & 14.02 & 15.12 & 17.74 \\
& SSIM$\uparrow$           & 0.739 & 0.803 & 0.572 & 0.604 & 0.436 & 0.497 & 0.608 \\
& LPIPS$\downarrow$        & 0.213 & 0.182 & 0.405 & 0.355 & 0.581 & 0.560 & 0.382 \\
\midrule
\multirow{3}{*}{SplaTAM \cite{keetha2023splatam}} 
& PSNR{[}dB{]}$\uparrow$  & 24.12 & 22.07 & 19.97 & 16.70 & 17.03 & 16.54 & 19.40 \\
& SSIM$\uparrow$           & \textbf{0.915} & \textbf{0.879} & 0.799 & 0.688 & 0.650 & 0.635 & 0.757 \\
& LPIPS$\downarrow$        & 0.101 & 0.163 & 0.205 & 0.287 & 0.339 & 0.353 & 0.241 \\
\midrule
\multirow{3}{*}{SC-GS \cite{scgs}} 
& PSNR{[}dB{]}$\uparrow$  & 27.01 & 21.45 & 18.93 & 20.99 & 19.89 & 16.44 & 20.78 \\
& SSIM$\uparrow$           & 0.900 & 0.686 & 0.529 & 0.762 & 0.590 & 0.475 & 0.657 \\
& LPIPS$\downarrow$        & 0.182 & 0.369 & 0.512 & 0.291 & 0.470 & 0.554 & 0.396 \\
\midrule
\multirow{3}{*}{4DGS-SLAM \cite{4dgs-slam}} 
& PSNR{[}dB{]}$\uparrow$  & 27.68 & 24.37 & 20.71 & 23.21 & 19.83 & 19.47 & 22.55 \\
& SSIM$\uparrow$           & 0.892 & 0.822 & 0.746 & 0.822 & 0.730 & 0.713 & 0.788 \\
& LPIPS$\downarrow$        & 0.116 & 0.179 & 0.265 & 0.192 & 0.281 & 0.340 & 0.229 \\
\midrule
\multirow{3}{*}{Ours} 
& PSNR{[}dB{]}$\uparrow$  & \textbf{29.70} & \textbf{27.49} & \textbf{26.04} & \textbf{27.64} & \textbf{24.60} & \textbf{23.83} & \textbf{26.55} \\
& SSIM$\uparrow$           & 0.905 & 0.860 & \textbf{0.812} & \textbf{0.852} & \textbf{0.788} & \textbf{0.769} & \textbf{0.831} \\
& LPIPS$\downarrow$        & \textbf{0.092} & \textbf{0.150} & \textbf{0.223} & \textbf{0.127} & \textbf{0.206} & \textbf{0.263} & \textbf{0.177} \\
\bottomrule
\end{tabular}
\label{tab:TUM_render}
\end{table*}

\smallskip\noindent\textbf{GMM opacity and rotation.}
To represent the temporal opacity and rotations smoothly without timestamp-specific storage,
we model both the opacity $\sigma_i(t)$ and rotation $\mathbf{q}_i(t)$ using a Gaussian mixture model over normalized time $\hat{t}\in[0,1]$. 
For each Gaussian we learn $K$ mixture components with learnable weights $w_{i,k}$, temporal means $\mu_{i,k}$ and scales $\tau_{i,k}$. 
The opacity coefficient is defined as:
\begin{equation}
    m_i(t) = 1 - \exp\!\left(-A_i \sum_{k=1}^K 
    w_{i,k} \, 
    \mathcal{N}(\hat{t}; \mu_{i,k}, \tau_{i,k}^2) \right),
\end{equation}
where $A_i>0$ is a learnable activation amplitude. 
The final time-varying opacity is $\sigma_i(t) = \sigma_i \cdot m_i(t)$.

Each component $k$ also stores a learnable control quaternion $\mathbf{q}_{i,k}$, and the Gaussian activations serve as blending weights to achieve smooth rotation:
\begin{equation}
    \mathbf{q}_i(t) = 
    \frac{
        \sum_{k=1}^K 
        w_{i,k} \, 
        \mathcal{N}(\hat{t}; \mu_{i,k}, \tau_{i,k}^2)
        \, \mathbf{q}_{i,k}
    }{
        \left\|
        \sum_{k=1}^K 
        w_{i,k} \, 
        \mathcal{N}(\hat{t}; \mu_{i,k}, \tau_{i,k}^2)
        \, \mathbf{q}_{i,k}
        \right\|
    }.
\end{equation}
This formulation produces continuous and smoothly varying rotations. 
We set $K=3$ in the experiments.

In our design, explicit keyframe positions are used by the optical-flow–assisted 4D mapping (\cf Sec.~\ref{sec:mapping}) to accelerate training. In addition, GMM-based temporal opacity and rotation ensures smooth temporal behavior with minimal model-size overhead.

\subsection{Tracking}

Given the previously computed dynamic mask $\mathcal{M}_{dy}$, we follow 4DGS-SLAM \cite{4dgs-slam} to optimize the camera pose based on $L_1$ loss between the static-3DGS-rendered color and depth maps and the inputs. Specifically, we first compute an opacity mask and the final valid mask as:
\begin{subequations}
\label{eq:mv_omo}
\begin{align}
& \mathcal{M}_o(\bm{u}) = \bm{\mathds{1}}\big(\hat{\mathbf{O}}(\mathbf{u}) \ge \alpha\big),\\
& \mathcal{M}_{v} = (\neg\mathcal{M}_{dy}) \cap \mathcal{M}_{o}.
\end{align}
\end{subequations}

\begin{figure*}[t]
  \centering
  \captionsetup[sub]{labelformat=empty}

  \subfloat{%
		\begin{minipage}[b]{0.25\linewidth}
			\includegraphics[width=1\linewidth]{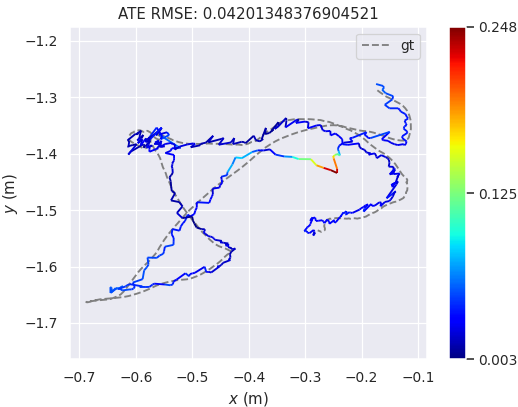}%
		\end{minipage}
	}
  \subfloat{%
		\begin{minipage}[b]{0.25\linewidth}
			\includegraphics[width=1\linewidth]{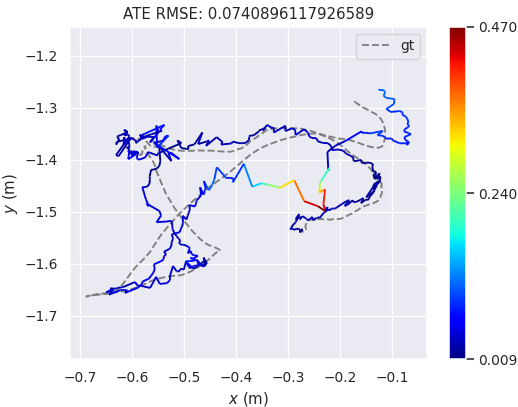}%
		\end{minipage}
	}
  \subfloat{%
		\begin{minipage}[b]{0.25\linewidth}
			\includegraphics[width=1\linewidth]{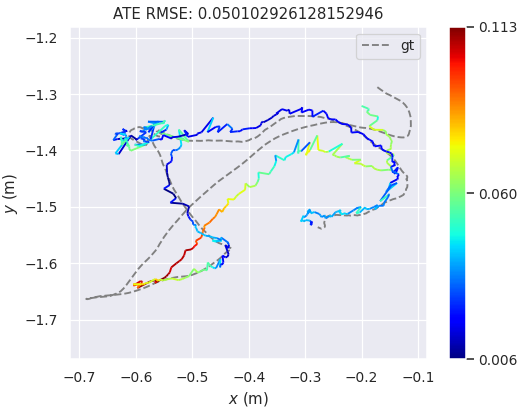}%
		\end{minipage}
	}
  \subfloat{%
		\begin{minipage}[b]{0.25\linewidth}
			\includegraphics[width=1\linewidth]{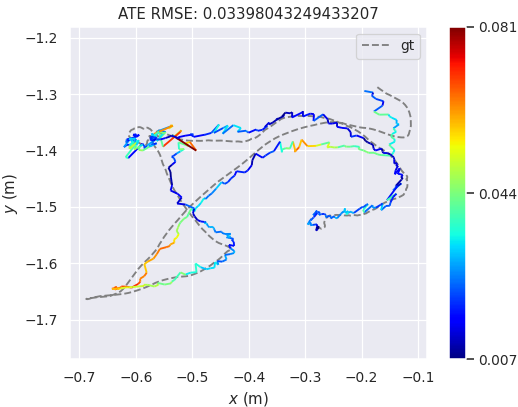}%
		\end{minipage}
	}\\
  \subfloat[(a) 4DGS-SLAM \cite{4dgs-slam}]{%
		\begin{minipage}[b]{0.25\linewidth}
			\includegraphics[width=1\linewidth]{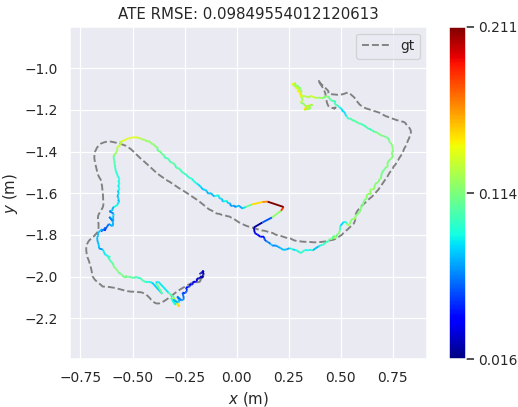}%
		\end{minipage}
	}
  \subfloat[(b) w/o Motion Decomp.]{%
		\begin{minipage}[b]{0.25\linewidth}
			\includegraphics[width=1\linewidth]{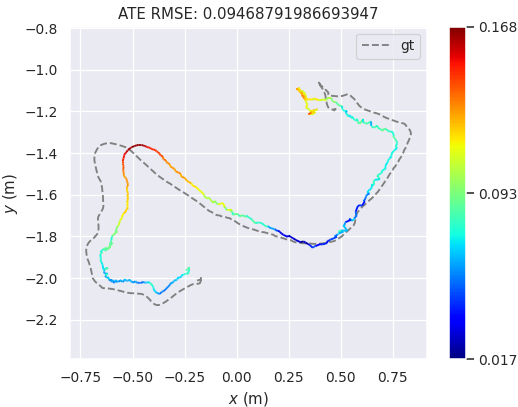}%
		\end{minipage}
	}
  \subfloat[(c) w/o Camera Init.]{%
		\begin{minipage}[b]{0.25\linewidth}
			\includegraphics[width=1\linewidth]{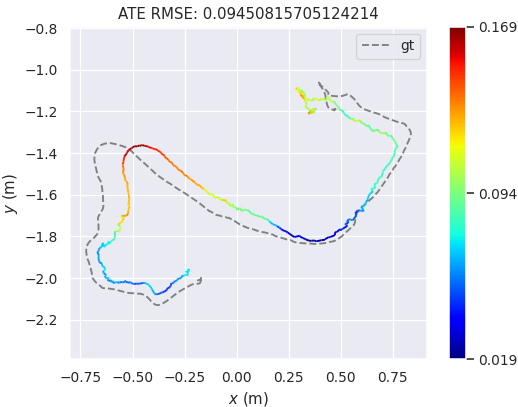}%
		\end{minipage}
	}
  \subfloat[(d) Ours]{%
		\begin{minipage}[b]{0.25\linewidth}
			\includegraphics[width=1\linewidth]{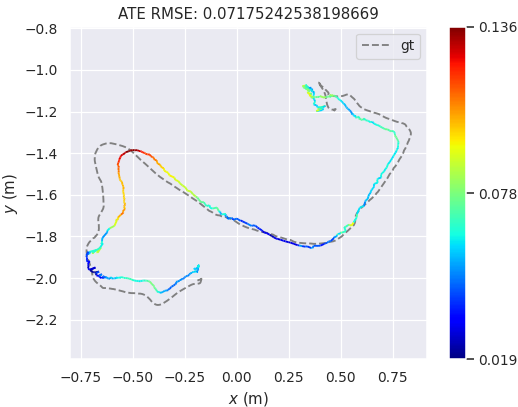}%
		\end{minipage}
	}

  \caption{\textbf{Tracking Results on BONN \cite{refusion} ballon2 (first row) and person\_tracking (second row) scenes.}}
  \label{fig:tracking}
\end{figure*}

\noindent 
The tracking loss is then computed as:
\begin{equation}
    \mathcal{L}_{track} = \frac{1}{|\mathcal{V}|}\sum_{\bm{u}\in\mathcal{V}}\mathcal{M}_{v}(\bm{u})\left(\lambda_1L_1(\hat{\mathbf{C}}(\bm{u}))+\lambda_2L_1(\hat{\mathbf{D}}(\bm{u}))\right),
\end{equation}
\noindent where $\mathcal{V}$ is the set of pixels with valid depths. Note that before tracking, the camera pose is initialized with the estimated camera motion from Sec. \ref{sec:motion} 
to form a coarse-to-fine pipeline for improved robustness (\textit{cf}. Figure \ref{fig:tracking}).

\begin{table*}[t]
\small
\centering
\renewcommand{\arraystretch}{1} 
\setlength{\tabcolsep}{6.1pt}       
\setlength{\aboverulesep}{0pt}
\setlength{\belowrulesep}{0pt}
\setlength{\extrarowheight}{0pt}
\caption{\textbf{Trajectory ATE RMSE [cm]$\downarrow$ on the BONN sequences.} 
Best results are highlighted in \textbf{bold}.}
\vspace{-3mm}
\begin{tabular}{l|cccccccccc}
\toprule
Method 
& ballon & ballon2 & ps\_track & ps\_track2 
& sync & sync2 & p\_no\_box & p\_no\_box2 & p\_no\_box3 & Avg. \\
\midrule
RoDyn-SLAM \cite{jiang2024rodynslam}      
& 7.9 & 11.5 & 14.5 & 13.8 & 1.3  & 1.4  & 4.9 & 6.2 & 10.2 & 7.9 \\
MonoGS \cite{monogs} 
& 29.6 & 22.1 & 54.5 & 36.9 & 68.5 & 0.56 & 71.5 & 10.7 & 3.6  & 33.1 \\
SplaTAM \cite{keetha2023splatam}          
& 32.9 & 30.4 & 77.8 & 116.7 & 59.5 & 66.7 & 91.9 & 18.5 & 17.1 & 56.8 \\
4DGS-SLAM \cite{4dgs-slam}                
& \textbf{2.5} & 4.2 & 9.8 & 11.2 & 0.95 & \textbf{0.64} & \textbf{1.8} & 1.7 & 2.6 & 3.9 \\
\midrule
Ours                                      
& 2.6 & \textbf{3.4} & \textbf{7.2} & \textbf{10.8} & \textbf{0.60} & \textbf{0.64} & 2.2 & \textbf{1.6} & \textbf{2.3} & \textbf{3.5} \\
\bottomrule
\end{tabular}
\label{BONN ATE}
\end{table*}

\begin{table*}[ht]
\small
\centering
\captionsetup[table*]{singlelinecheck=off}
\renewcommand{\arraystretch}{1} 
\setlength{\tabcolsep}{3.6pt}       
\setlength{\aboverulesep}{0pt}
\setlength{\belowrulesep}{0pt}
\setlength{\extrarowheight}{0pt}
\caption{\textbf{Quantitative results on the BONN sequences.}
Best scores are shown in \textbf{bold}. ``-'' indicates reconstruction failure.}
\vspace{-3mm}
\begin{tabular}{l|c|cccccccccc}
\toprule
Method & Metric 
& ballon & ballon2 & ps\_track & ps\_track2 & sync & sync2 
& p\_no\_box & p\_no\_box2 & p\_no\_box3 & Avg. \\
\midrule
\multirow{3}{*}{MonoGS \cite{monogs}} 
& PSNR{[}dB{]}$\uparrow$  & 21.35 & 20.22 & 20.53 & 20.09 & 22.03 & 20.55 & 20.76 & 19.38 & 24.81 & 21.06 \\
& SSIM$\uparrow$           & 0.803 & 0.758 & 0.779 & 0.718 & 0.766 & 0.841 & 0.748 & 0.753 & 0.857 & 0.780 \\
& LPIPS$\downarrow$        & 0.316 & 0.354 & 0.408 & 0.426 & 0.328 & 0.521 & 0.428 & 0.372 & 0.243 & 0.342 \\
\midrule
\multirow{3}{*}{SplaTAM \cite{keetha2023splatam}} 
& PSNR{[}dB{]}$\uparrow$  & 19.65 & 17.67 & 18.30 & 15.57 & 19.33 & 19.67 & 20.81 & 21.69 & 21.41 & 19.34 \\
& SSIM$\uparrow$           & 0.781 & 0.702 & 0.670 & 0.606 & 0.776 & 0.730 & 0.824 & 0.852 & 0.873 & 0.757 \\
& LPIPS$\downarrow$        & 0.211 & 0.280 & 0.283 & 0.331 & 0.227 & 0.258 & 0.191 & 0.165 & 0.152 & 0.233 \\
\midrule
\multirow{3}{*}{SC-GS \cite{scgs}} 
& PSNR{[}dB{]}$\uparrow$  & 22.30 & 21.38 & -    & -    & 23.62 & 22.74 & 20.60 & 21.55 & 19.24 & 21.63 \\
& SSIM$\uparrow$           & 0.737 & 0.708 & -    & -    & 0.788 & 0.801 & 0.688 & 0.722 & 0.628 & 0.724 \\
& LPIPS$\downarrow$        & 0.448 & 0.450 & -    & -    & 0.427 & 0.359 & 0.515 & 0.491 & 0.539 & 0.461 \\
\midrule
\multirow{3}{*}{4DGS-SLAM \cite{4dgs-slam}} 
& PSNR{[}dB{]}$\uparrow$  & 26.19 & 22.91 & 21.78 & 20.65 & 24.37 & 25.12 & 23.14 & 24.28 & 25.88 & 23.81 \\
& SSIM$\uparrow$           & 0.875 & 0.839 & 0.832 & 0.820 & 0.833 & 0.892 & 0.845 & 0.873 & 0.886 & 0.855 \\
& LPIPS$\downarrow$        & 0.235 & 0.269 & 0.289 & 0.294 & 0.230 & 0.173 & 0.239 & 0.224 & 0.207 & 0.240 \\
\midrule
\multirow{3}{*}{Ours} 
& PSNR{[}dB{]}$\uparrow$  & \textbf{30.30} & \textbf{28.36} & \textbf{29.18} & \textbf{28.94} 
& \textbf{29.85} & \textbf{31.26} & \textbf{29.25} & \textbf{30.03} & \textbf{30.22} & \textbf{29.71} \\
& SSIM$\uparrow$           & \textbf{0.897} & \textbf{0.877} & \textbf{0.879} & \textbf{0.882} 
& \textbf{0.874} & \textbf{0.924} & \textbf{0.883} & \textbf{0.896} & \textbf{0.900} & \textbf{0.890} \\
& LPIPS$\downarrow$        & \textbf{0.193} & \textbf{0.204} & \textbf{0.228} & \textbf{0.211} 
& \textbf{0.165} & \textbf{0.140} & \textbf{0.204} & \textbf{0.205} & \textbf{0.187} & \textbf{0.193} \\
\bottomrule
\end{tabular}
\label{tab:BONN_render}
\end{table*}

\begin{figure*}[htbp]
	\centering
    \captionsetup[subfloat]{labelformat=empty}
	\subfloat{%
        \hspace{-5mm}%
        \rotatebox{90}{\scriptsize{~~~~~~~~~~~~~~\textbf{fr3/walk\_static}}}
		\begin{minipage}[b]{0.25\linewidth}
			\includegraphics[width=1\linewidth]{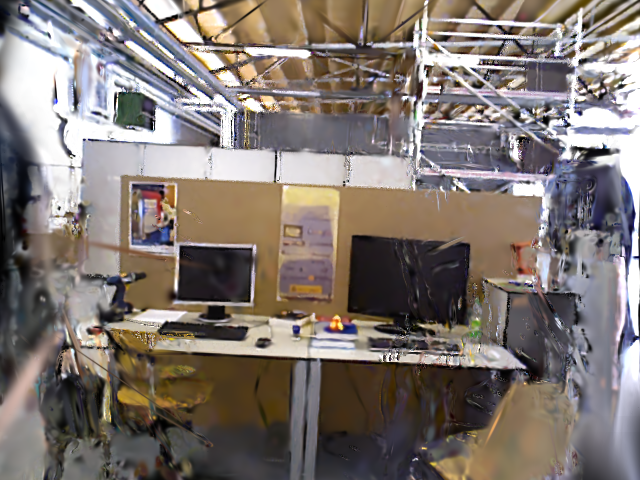}%
		\end{minipage}
	}
	\subfloat{%
		\begin{minipage}[b]{0.25\linewidth}
			\includegraphics[width=1\linewidth]{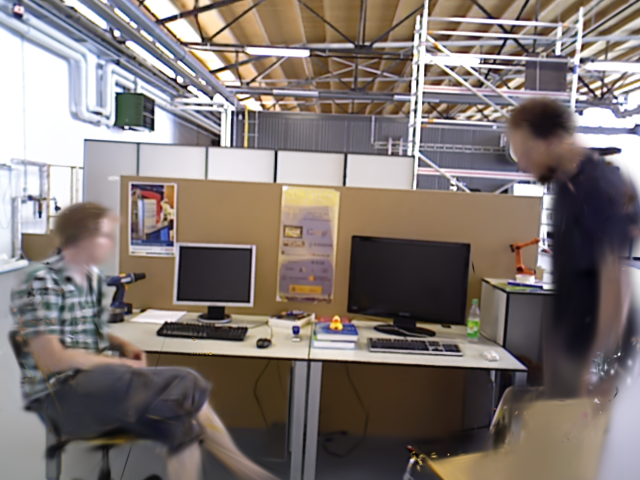}%
		\end{minipage}
	}
	\subfloat{%
		\begin{minipage}[b]{0.25\linewidth}
			\includegraphics[width=1\linewidth]{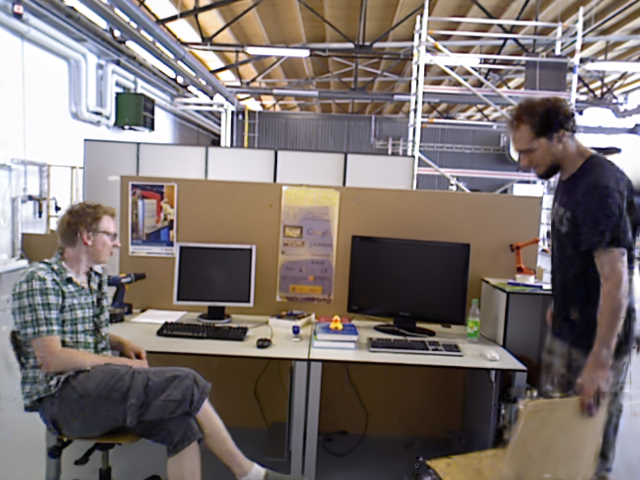}%
		\end{minipage}
	}
    \subfloat{%
		\begin{minipage}[b]{0.25\linewidth}
			\includegraphics[width=1\linewidth]{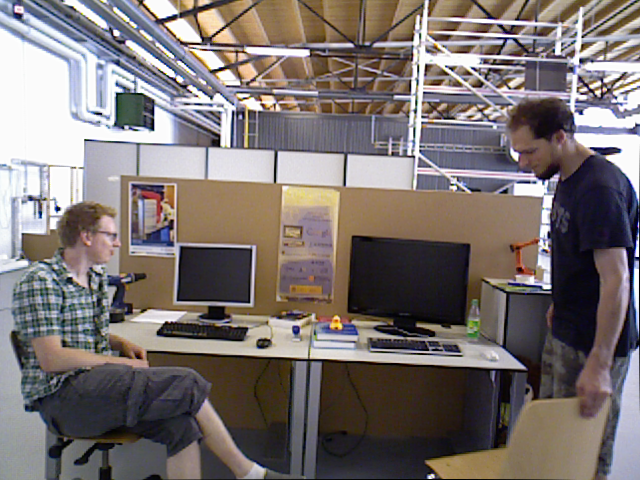}%
		\end{minipage}
	}\\
	\vspace{0.5mm}
    \subfloat{%
        \hspace{-5mm}%
         \rotatebox{90}{\scriptsize{~~~~~~~~~~~~~~~\textbf{fr3/walk\_xyz}}}
		\begin{minipage}[b]{0.25\linewidth}
			\includegraphics[width=1\linewidth]{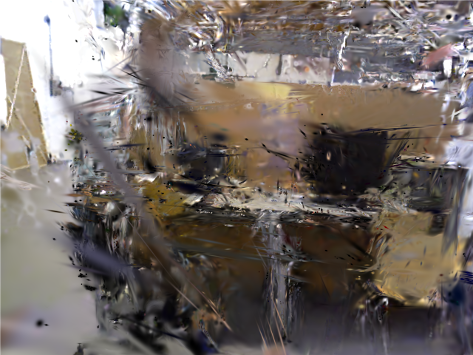}%
		\end{minipage}
	}
	\subfloat{%
		\begin{minipage}[b]{0.25\linewidth}
			\includegraphics[width=1\linewidth]{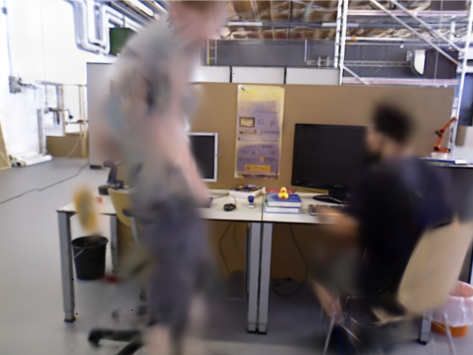}%
		\end{minipage}
	}
	\subfloat{%
		\begin{minipage}[b]{0.25\linewidth}
			\includegraphics[width=1\linewidth]{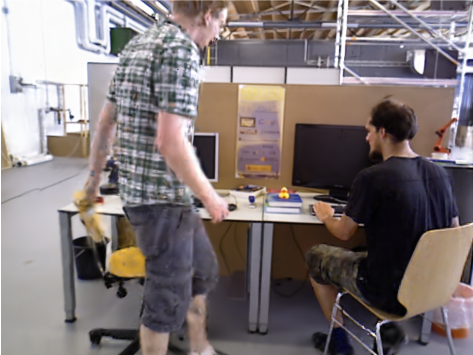}%
		\end{minipage}
	}
    \subfloat{%
		\begin{minipage}[b]{0.25\linewidth}
			\includegraphics[width=1\linewidth]{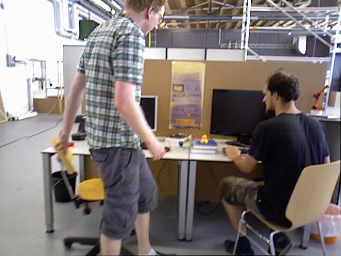}%
		\end{minipage}
	}\\
	\vspace{0.5mm}
    \subfloat{%
        \hspace{-5mm}%
        \rotatebox{90}{\scriptsize{~~~~~~~~~~~~~~\textbf{bonn\_ballon2}}}
		\begin{minipage}[b]{0.25\linewidth}
			\includegraphics[width=1\linewidth]{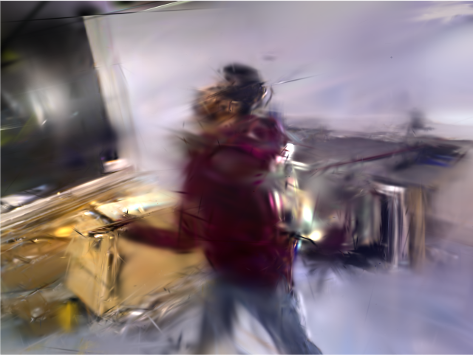}%
		\end{minipage}
	}
	\subfloat{%
		\begin{minipage}[b]{0.25\linewidth}
			\includegraphics[width=1\linewidth]{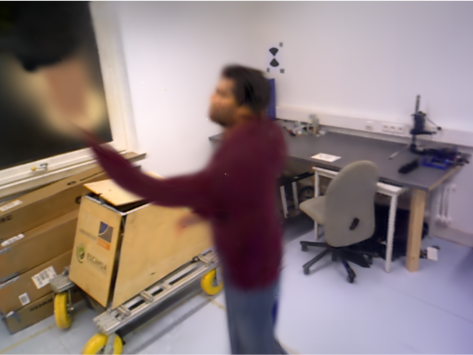}%
		\end{minipage}
	}
	\subfloat{%
		\begin{minipage}[b]{0.25\linewidth}
			\includegraphics[width=1\linewidth]{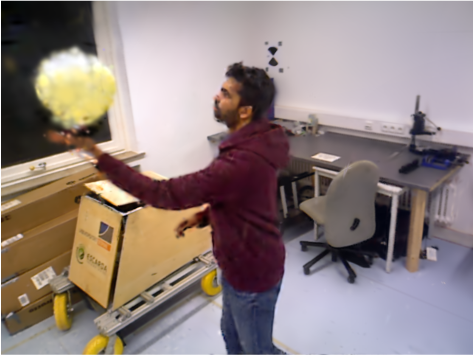}%
		\end{minipage}
	}
    \subfloat{%
		\begin{minipage}[b]{0.25\linewidth}
			\includegraphics[width=1\linewidth]{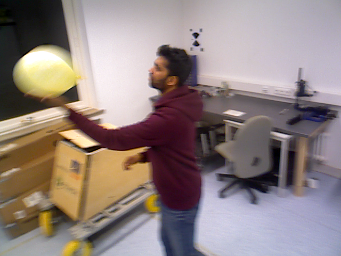}%
		\end{minipage}
	}\\
	\vspace{0.5mm}
    \subfloat[MonoGS \cite{monogs}]{%
        \hspace{-5mm}%
        \rotatebox{90}{\scriptsize{~~~~~~~~~~~~~~\textbf{bonn\_ps\_trk2}}}
		\begin{minipage}[b]{0.25\linewidth}
			\includegraphics[width=1\linewidth]{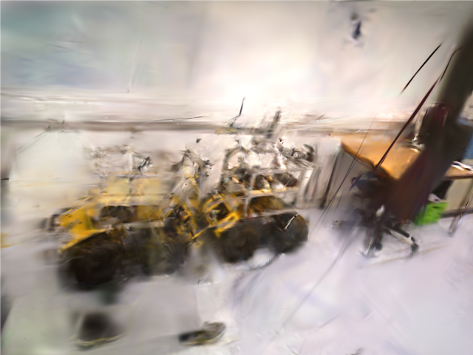}%
		\end{minipage}
	}
	\subfloat[4DGS-SLAM \cite{4dgs-slam}]{%
		\begin{minipage}[b]{0.25\linewidth}
			\includegraphics[width=1\linewidth]{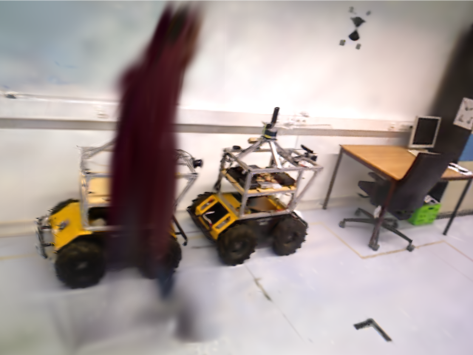}%
		\end{minipage}
	}
	\subfloat[Ours]{%
		\begin{minipage}[b]{0.25\linewidth}
			\includegraphics[width=1\linewidth]{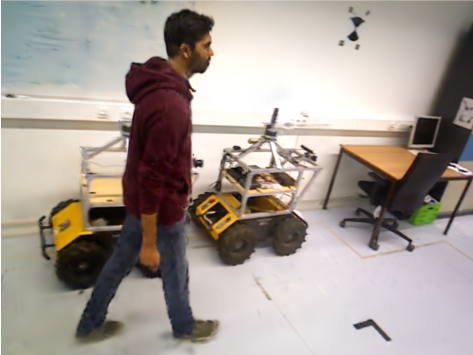}%
		\end{minipage}
	}
    \subfloat[Ground Truth]{%
		\begin{minipage}[b]{0.25\linewidth}
			\includegraphics[width=1\linewidth]{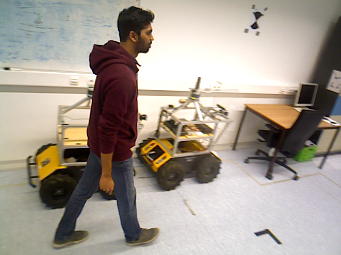}%
		\end{minipage}
	}\\

	\caption{\textbf{Qualitative Renderings on \textit{non-keyframes} in TUM RGB-D \cite{tumbenchmark} (first two rows) and BONN \cite{refusion} (last two rows) datasets.}}
    \vspace{-3mm}
	\label{fig:render}
\end{figure*}

\subsection{Optical Flow-Guided 4D Mapping}
\label{sec:mapping}
We introduce a scene-flow–based Gaussian propagation module and an adaptive Gaussian insertion module, to accelerate training of dynamic Gaussians for online SLAM and efficiently handle newly appearing dynamic objects. Both modules explicitly operate on Gaussian positions.

\smallskip
\noindent\textbf{Scene Flow Gaussian Propagation.} 
Before mapping keyframe $k$, we maintain the current set of dynamic Gaussians $\bm{\mathcal{G}}^{t_{k-1}}$ with 3D centers $\{\mathbf{x}_i^{k-1}\}$. We then initialize the dynamic Gaussians at keyframe $k$ by propagating these centers using the prior scene flow.
Let $\mathbf{K} \in \mathbb{R}^{3\times 3}$ be the camera intrinsic matrix, and 
$(\mathbf{R}_k, \mathbf{t}_k)$ denote the optimized rotation and translation of keyframe-$k$ camera, 
we get the projection matrix 
$\mathbf{P}_k = \mathbf{K}[\mathbf{R}_k \mid \mathbf{t}_k]$.

Each Gaussian center is first projected into the image of keyframe $k-1$: 
$
    \bar{\mathbf{u}}_i^{\,k-1}
    = \mathbf{P}_{k-1}
    \begin{bmatrix}
    \mathbf{x}_i^{k-1} \\
    1
    \end{bmatrix},
    \quad
    \mathbf{u}_i^{k-1} = \Pi(\bar{\mathbf{u}}_i^{\,k-1}),
$
where $\Pi(\cdot)$ performs homogeneous normalization. 
With the predicted optical flow $\mathbf{F}^{t_{k-1}, t_k}$, we propagate these projections to keyframe $k$ as
$
    \mathbf{u}_i^{k}
    = \mathbf{u}_i^{k-1}
    + \mathbf{F}^{t_{k-1}, t_k}(\mathbf{u}_i^{k-1}).
$ 
Subsequently, we obtain a coarse estimate of the 3D deformation from time $k-1$ to $k$ by unprojection:
\begin{equation}
\Delta \mathbf{x}_i^{k}
=
\mathbf{R}_{k}^{\!\top}
\!\left(
\mathrm{D}_i^{k}\,\mathbf{K}^{-1}\bar{\mathbf{u}}_i^k
- \mathbf{t}_{k}
\right)
-
\mathbf{x}_i^{k-1},
\end{equation}
where $\bar{\mathbf{u}}_i^k = [\mathbf{u}_i^{k}, 1]^\top$ is the homogeneous pixel location. 

We mitigate noise in the flow estimation and enforce local rigidity by regularizing $\Delta\mathbf{x}_i^k$ with KNN smoothing:
\begin{gather}
    \Delta \widehat{\mathbf{x}}_i^k
    =
    \sum_{j\in\mathcal{N}(i)}
    w_{ij}^{knn}\,
    \Delta \mathbf{x}_j^k, \\
    w_{ij}^{knn}
    =
    \frac{\mathcal{N}\!\left(
    \left\Vert\mathbf{x}_j^{k-1}-
    \mathbf{x}_i^{k-1}\right\Vert_2
    \,;\,
    0,
    \tau_{knn}^2
    \right)}{\sum\limits_{l\in\mathcal{N}(i)}\mathcal{N}\!\left(
    \left\Vert\mathbf{x}_l^{k-1}-
    \mathbf{x}_i^{k-1}\right\Vert_2
    \,;\,
    0,
    \tau_{knn}^2
    \right)},
\end{gather}
where $\mathcal{N}(i)$ is the nearest neighbors set with a radius of the $i$-th dynamic Gaussian. 
Consequently, the corresponding dynamic Gaussian center at keyframe $k$ is initialized as $
\mathbf{x}_i^{k}
=
\mathbf{x}_i^{k-1}
+
\Delta \widehat{\mathbf{x}}_i^k$.
This yields motion-aware and spatially consistent Gaussian initialization, which substantially accelerates convergence in the subsequent 4D optimization.

\smallskip
\noindent\textbf{Adaptive Gaussian Insertion.} Although the Gaussian propagation updates existing dynamic Gaussians to a better initialization at the new keyframe, new dynamic objects or previously occluded regions of dynamic objects may appear in keyframe $k$ that was absent in keyframe $k-1$. To ensure complete coverage and fast convergence on such dynamic regions, we initialize new Gaussians based on optical flow back-tracking.

Let $\mathcal{M}_{dy}^k$ and $\mathcal{M}^{k-1}_{dy}$ denote the binary motion masks of keyframes $k$ and $k-1$, respectively. We first warp the current mask into the previous keyframe using the predicted backward flow $\mathbf{F}^{t_{k}, t_{k-1}}$:
$
\mathbf{u}_p^{k-1}
=
\mathbf{u}_p^{k}
+
\mathbf{F}^{t_{k}, t_{k-1}}(\mathbf{u}_p^k),
$
for each pixel $\mathbf{u}_p^k\in\mathcal{M}_{dy}^{t_k}$.

Pixels of new dynamic regions are identified by:
\begin{equation}
\mathcal{M}_{\text{insert}}^{t_{k}}
=
\left\{
\mathbf{u}_p^k \in \mathcal{M}_{dy}^{t_k} \;\big|\;
\mathbf{u}_p^{k-1} \notin \mathcal{M}^{t_{k-1}}_{dy}
\right\}.
\end{equation}
We randomly sample these newly activated dynamic pixels with a density factor $1/D_{\text{init}}$ and initialize new dynamic Gaussians by unprojecting sampled pixels.

\smallskip
\noindent\textbf{Optimization.} 
We conduct a fast training step after the Gaussian propagation and adaptive gaussian insertion steps.
We follow 4DGS-SLAM \cite{4dgs-slam} to train on a sliding window of keyframes and $2$ random previous keyframes, but for much fewer iterations (which we set to $50$ in the experiments). 
Finally, the mapping loss is:
\begin{equation}
    \mathcal{L}_{map} = \lambda_1\mathcal{L}_{c}+\lambda_2\mathcal{L}_{d}+\lambda_{f}\mathcal{L}_{f}+\lambda_{m}\mathcal{L}_{m}+\lambda_{iso}\mathcal{L}_{iso},
\end{equation}
where $\mathcal{L}_{c}$, $\mathcal{L}_{d}$, $\mathcal{L}_{f}$, $\mathcal{L}_{iso}$ 
are the color, depth, flow, and isotropic loss, respectively. $\mathcal{L}_m$ is a binary mask loss which constrains rendered dynamic Gaussians' alpha map to be consistent with motion mask. $\lambda_{f}$, $\lambda_m$ and $\lambda_{iso}$ are the corresponding weights. The flow loss and isotropic loss are adopted from 4DGS-SLAM \cite{4dgs-slam}.

\section{Experiments}
\subsection{Experimental Settings}

\noindent\textbf{Datasets.} We evaluate our method in two real-world datasets: TUM RGB-D dataset \cite{tumbenchmark} and BONN dataset \cite{refusion}. We mainly evaluate on the dynamic sequences following 4DGS-SLAM \cite{4dgs-slam} to form a direct comparison.

\smallskip
\noindent\textbf{Implementation Details.} We ran our experiments on single NVIDIA RTX A6000 GPU. For tracking, we follow 4DGS-SLAM \cite{4dgs-slam} to use $100$ and $200$ maximum iterations on TUM RGB-D \cite{tumbenchmark} and BONN \cite{refusion} datasets, respectively. For mapping, we train for $50$ iterations and use window size of $8$. 
We only apply flow loss in the last $25$ iterations to speed up each mapping step. 
We follow the keyframe selection strategy of 4DGS-SLAM, which selects keyframes based on motion mask differences and at least selects 1 keyframe in every 5 frames. After online training, we also conduct $1500$ iterations of color refinement following 4DGS-SLAM \cite{4dgs-slam} and MonoGS \cite{monogs}. For other detailed hyperparameters, 
please refer to our supplementary material. 

\smallskip
\noindent\textbf{Baseline Methods.} We mainly compare with 3DGS-SLAM methods without explicit loop closure. For static SLAM, we compare with MonoGS \cite{monogs}, SplaTAM \cite{keetha2023splatam}. For dynamic reconstruction and SLAM, we compare with SC-GS \cite{scgs} and 4DGS-SLAM \cite{4dgs-slam}. For 4DGS-SLAM, we report our reproduced results using their officially released code.

\smallskip
\noindent\textbf{Metrics.} For camera tracking, we follow the standard SLAM methods to evaluate the Root Mean Square Error (RMSE) of the Absolute Trajectory error (ATE) across all frames. For photometric rendering quality, we report the PSNR, SSIM, and LPIPS on all frames.



\subsection{Main Results}
\noindent\textbf{Camera Tracking.} The camera tracking results on TUM RGB-D \cite{tumbenchmark} and BONN \cite{refusion} are reported in Tables \ref{tab:tum_ate} and \ref{BONN ATE}. Our method achieves more accurate camera trajectories on both datasets. Notably, 4DGS-SLAM \cite{4dgs-slam} uses prolonged mapping ($200$ iterations) to reconstruct the scene and refine camera poses within a pose window of length $3$, whereas we use far fewer mapping iterations while still obtaining better tracking accuracy. A qualitative comparison is shown in Figure \ref{fig:tracking}. Without the motion decomposition module (\texttt{w/o Motion Decomp.}), our method performs worse than 4DGS-SLAM due to limited camera pose refinement. Adding category-agnostic motion masking (\texttt{w/o Camera Init}) markedly improves the results on \texttt{ballon2}, which contains category-agnostic dynamic objects. Furthermore, flow-guided camera initialization improves tracking accuracy on both scenes and mitigates camera trajectory drift.

\smallskip
\noindent\textbf{Photometric Renderings.} As shown in Tables \ref{tab:TUM_render} and \ref{tab:BONN_render}, our method achieves substantially higher photometric quality than the baselines. Compared with 4DGS-SLAM~\cite{4dgs-slam}, the main advantage comes from our hybrid dynamic Gaussian representation, supported by scene-flow based propagation and adaptive insertion for more effective dynamic modeling. Qualitative results are shown in Figure \ref{fig:render}. 4DGS-SLAM underperforms in highly dynamic scenes, \textit{e.g.} when people leave and re-enter the view (\texttt{fr3/walking\_xyz}), because it requires a handcrafted dynamic start time to assign dynamic Gaussians and cannot handle such complex dynamics well. In contrast, our method can adaptively insert new dynamic Gaussians without scene-specific hyperparameters. The BONN \cite{refusion} results also show that our method preserves more details in fast-moving scenes and reconstructs category-agnostic dynamic objects. We further visualize 2D tracking together with dynamic reconstruction in Figure \ref{teaser} by sampling 2D points at the start time and incrementally warping them with the rendered Gaussian flow maps. This shows that the learned Gaussian flow trajectories are spatially and temporally coherent.

\begin{table}[t]
\small
\centering
\renewcommand{\arraystretch}{1} 
\setlength{\tabcolsep}{1.3mm}     
\setlength{\aboverulesep}{0pt}
\setlength{\belowrulesep}{0pt}
\setlength{\extrarowheight}{0pt}
\vspace{-3mm}
\caption{\textbf{Running time (ms) analysis.}}
\vspace{-3mm}
\begin{tabular}{l|ccc|c}
\toprule
Method & Dynamic Seg. & Tracking & Mapping & FPS \\
\midrule
MonoGS \cite{monogs}       & -   & 476   & 557    & 1.93 \\
\midrule
4DGS-SLAM \cite{4dgs-slam} & \textbf{16} & 445 & 110562 & 0.04 \\
\textbf{Ours}              & 68  & \textbf{427} & \textbf{6285} & \textbf{0.50} \\
\bottomrule
\end{tabular}%
\label{tab:runtime}
\end{table}

\smallskip
\noindent\textbf{Runtime Analysis.} The runtime analysis is reported in Table \ref{tab:runtime}. 4DGS-SLAM \cite{4dgs-slam} has an extremely slow mapping speed due to its computationally expensive design. In each mapping step, it runs $100$ iterations to train the deformation MLPs and another $100$ iterations to jointly optimize the Gaussian parameters, which substantially reduces FPS. In contrast, our method benefits from explicit dynamic Gaussian centers guided by optical flow, leading to much faster mapping. For dynamic segmentation, our camera-induced motion segmentation takes around $52$ ms per frame and plays an indispensable role in our method (\textit{cf}. Table \ref{tab:ablation}, Figure \ref{fig:tracking}).

\subsection{Ablation Study}
The quantitative ablation results are shown in Table \ref{tab:ablation}. Our camera-induced motion decomposition module improves both camera tracking accuracy and rendering quality, especially on BONN \cite{refusion}, which contains category-agnostic dynamic objects and fast-moving scenes and cameras. The flow propagation and adaptive insertion modules contribute substantially to reconstruction quality in fast-moving scenes (\textit{e.g.} \texttt{ballon2}) and scenes with re-occurring dynamic objects (\textit{e.g.} \texttt{fr3/walk\_xyz}). Our GMM-based temporal opacity and rotation further improve representation capability in complex dynamic scenes, while the KNN smoothing strategy enhances the local rigidity of propagated Gaussians for fast mapping.

\begin{table}[t]
\small
\centering
\renewcommand{\arraystretch}{1} 
\setlength{\tabcolsep}{0.8mm}
\setlength{\aboverulesep}{0pt}
\setlength{\belowrulesep}{0pt}
\setlength{\extrarowheight}{0pt}
\vspace{-3mm}
\caption{\textbf{Ablation study.}}
\vspace{-3mm}
\begin{tabular}{l|cc|cc}
\toprule
\multirow{2}{*}{Method}
& \multicolumn{2}{c|}{fr3/walk\_xyz}
& \multicolumn{2}{c}{ballon2} \\
& ATE [cm]$\downarrow$ & PSNR$\uparrow$
& ATE [cm]$\downarrow$ & PSNR$\uparrow$ \\
\midrule
\textit{w/o} Motion Decomp.     & 2.7 & 24.40 & 7.4 & 27.59 \\
\textit{w/o} Flow Propagate     & 2.6 & 23.91 & 4.2 & 26.86 \\
\textit{w/o} Adaptive Insert    & 3.4 & 23.53 & 3.9 & 27.93 \\
\textit{w/o} GMM                & 2.7 & 24.04 & 3.7 & 27.91 \\
\textit{w/o} KNN smooth         & \textbf{2.5} & 24.47 & 3.5 & 28.14 \\
\midrule
\textbf{Ours}                   & \textbf{2.5} & \textbf{24.60} & \textbf{3.4} & \textbf{28.36} \\
\bottomrule
\end{tabular}%
\label{tab:ablation}
\end{table}

\section{Conclusion}
In this paper, we proposed Flow4DGS-SLAM, a novel dynamic SLAM system designed to jointly reconstruct 
static and dynamic regions. We first propose a camera-induced motion decomposition module, which leverages depth and prior optical flow to not only generate category-agnostic motion mask but also provide 
better camera pose initialization. 
We accelerate the training of dynamic Gaussians with a hybrid dynamic Gaussian representation combined with optical flow-guided 4D mapping. 
The empirical results demonstrate 
the significance of our Flow4DGS-SLAM in dynamic 3DGS SLAM, showing superior performance on camera tracking, photometric rendering quality, and training efficiency.


\vspace{-5pt}
\paragraph{Acknowledgment.} This research / project is supported by the National Research Foundation (NRF) Singapore, under
its NRF-Investigatorship Programme (Award ID. NRF-NRFI09-0008), and the Tier 2 grant MOET2EP20124-0015 from the Singapore Ministry of Education.




{
    \small
    \bibliographystyle{ieeenat_fullname}
    \bibliography{main}
}


\end{document}